\pdfoutput=1

\documentclass[11pt]{article}

\usepackage[]{emnlp2021}

\usepackage{times}
\usepackage{latexsym}

\usepackage{graphicx}
\usepackage{float}
\usepackage{booktabs}
\usepackage{multirow}

\usepackage[T1]{fontenc}

\usepackage[utf8]{inputenc}

\usepackage{microtype}

\title{Zero-Shot Dialogue State Tracking via Cross-Task Transfer}

\author{Zhaojiang Lin$^1$\thanks{\quad Work done during internship at Facebook}$^*$, Bing Liu$^2$, Andrea Madotto$^1$ $^*$, Seungwhan Moon$^2$, Paul Crook$^2$,  \\ \textbf{Zhenpeng Zhou$^2$, Zhiguang Wang$^2$, Zhou Yu$^3$, Eunjoon Cho$^2$, Rajen Subba$^2$,} \\ \textbf{Pascale Fung$^1$}\\
  $^1$The Hong Kong University of Science and Technology \\
  $^2$Facebook, $^3$Columbia University \\
  \texttt{zlinao@ust.hk}, \texttt{bingl@fb.com}\\

  }

\begin{document}
\maketitle
\begin{abstract}
Zero-shot transfer learning for dialogue state tracking (DST) enables us to handle a variety of task-oriented dialogue domains without the expense of collecting in-domain data.
In this work, we propose to transfer the \textit{cross-task} knowledge from general question answering (QA) corpora for the zero-shot DST task. Specifically, we propose TransferQA, a transferable generative QA model that seamlessly combines extractive QA and multi-choice QA via a text-to-text transformer framework, and tracks both categorical slots and non-categorical slots in DST. In addition, we introduce two effective ways to construct unanswerable questions, namely, \textit{negative question sampling} and \textit{context truncation}, which enable our model to handle \textit{``none''} value slots in the zero-shot DST setting. The extensive experiments show that our approaches substantially improve the existing zero-shot and few-shot results on MultiWoz. Moreover, compared to the fully trained baseline on the Schema-Guided Dialogue dataset, our approach shows better generalization ability in unseen domains.
\end{abstract}

\section{Introduction}
Virtual assistants are designed to help users perform daily activities, such as travel planning, online shopping and restaurant booking. Dialogue state tracking (DST), as an essential component of these task-oriented dialogue systems, tracks users' requirements throughout multi-turn conversations as dialogue states, which are typically in the form of a list of slot-value pairs. Training a DST model often requires extensive annotated dialogue data. These data are often collected via a Wizard-of-Oz (Woz)~\cite{kelley1984iterative} setting, where two workers converse with each other and annotate the dialogue states of each utterance~\cite{wen2017network,budzianowski2018multiwoz,moon2020situated}, or with a Machines Talking To Machines (M2M) framework~\cite{shah2018building}, where dialogues are synthesized via the system and user  simulators~\cite{campagna2020zero,rastogi2020towards,lin2021bitod}.
However, both of these approaches have inherent challenges when scaling to large datasets.
For example, the data collection process in a Woz setting incurs expensive and time-consuming manual annotations, while M2M requires exhaustive hand-crafted rules for covering various dialogue scenarios.

\begin{figure*}[t]
    \centering
    \includegraphics[width=0.8\linewidth]{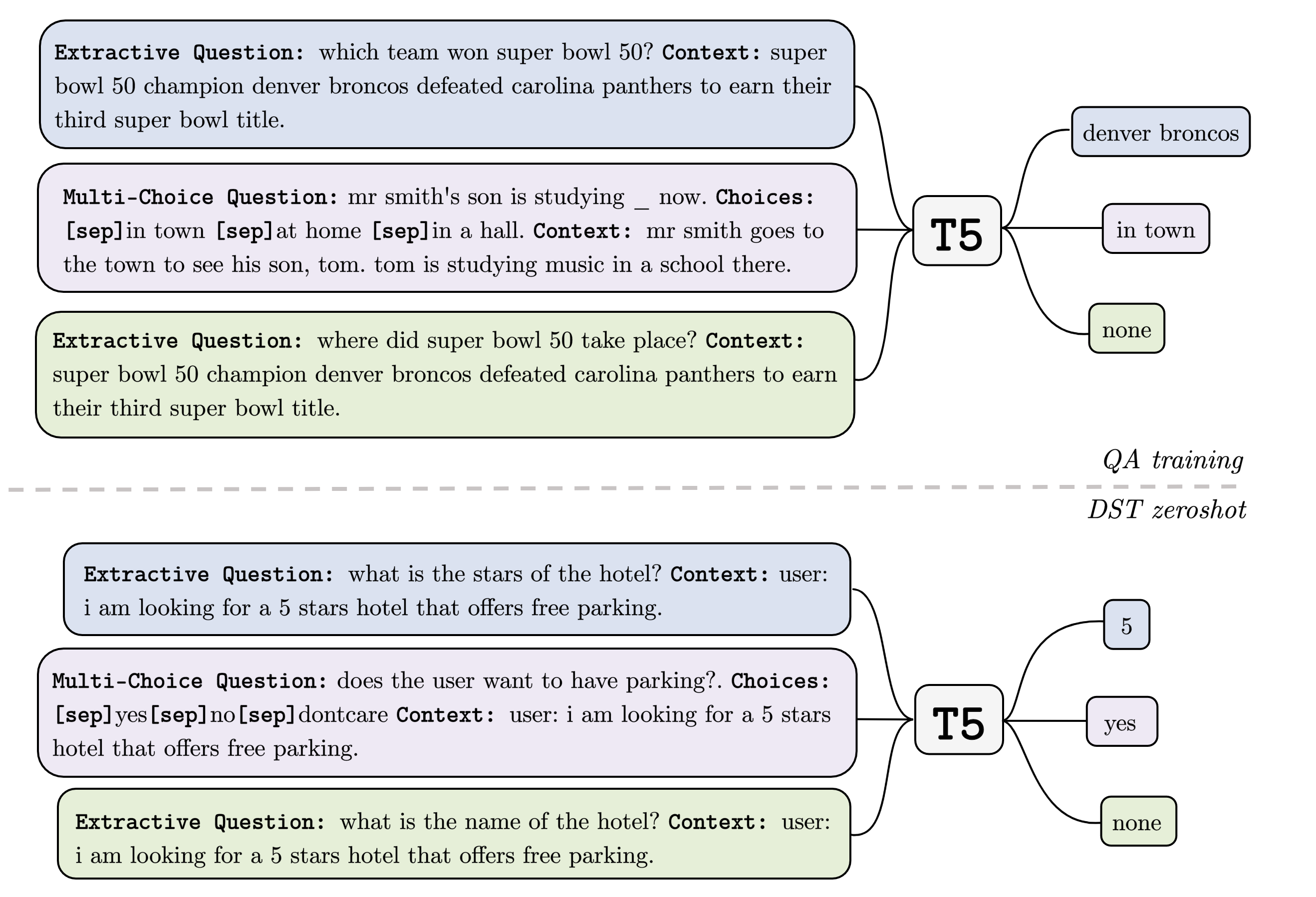}
    \caption{A high-level representation of the cross-task transfer for zero-shot DST (best viewed in color). During the QA training phase (top figure), the unified generative model (T5) is pre-trained on QA pairs of extractive questions (blue), multiple-choice questions (purple), and negative extractive questions (green).
    At inference time for zero-shot DST (bottom figure), the model predicts slot values as answers for synthetically formulated extractive questions (for non-categorical slots) and multiple-choice questions (for categorical slots). Note that the negative QA training allows for the model to effectively handle ``\textit{none}" values for unanswerable questions.}
    \label{fig:mainfigure}
\end{figure*}

In industrial applications, virtual assistants are required to add new services (domains) frequently based on user’s needs, but collecting extensive data for every new domain is costly and inefficient.
Therefore, performing zero-shot prediction of dialogue states is becoming increasingly important since it does not require the expense of data acquisition. There are mainly two lines of work in the zero-shot transfer learning problem. The first is cross-domain transfer learning~\cite{wu2019transferable,kumar2020ma,rastogi2020towards,lin2021leveraging}, where the models are first trained on several domains, then zero-shot to new domains. However, these methods rely on a considerable amount of DST data to cover a broad range of slot types, and it is still challenging for the models to handle new slot types in the unseen domain. The second line of work leverages machine reading question answering (QA) data to facilitate the low-resource DST (i.e., cross-task transfer)~\cite{gao2020machine}. However, the method of \citet{gao2020machine} relies on two independent QA models, i.e., a span extraction model for non-categorical slots and a classification model for categorical slots, which hinders the knowledge sharing from the different types of QA datasets. Furthermore, unanswerable questions are not considered during their QA training phase. Therefore, in a zero-shot DST setting, the model proposed by ~\citet{gao2020machine} is not able to handle \textit{``none''} value slots (e.g., unmentioned slots) that present in the dialogue state.

In this paper, to address the above challenges, we propose TransferQA, a unified generative QA model that seamlessly combines extractive QA and multi-choice QA via a text-to-text transformer framework~\cite{raffel2020exploring,khashabi2020unifiedqa}. Such design not only allows the model to leverage both extractive and multi-choice QA datasets, but also provides a simple unified text-to-text interface for tracking both categorical slots and non-categorical slots. 
To handle the \textit{``none''} value slots in a zero-shot DST setting, we introduce two effective ways to construct unanswerable questions, namely, \textit{negative question sampling} and \textit{context truncation}, which simulate the out-of-domain slots and in-domain unmentioned slots in multi-domain DST. We evaluate our approach on two large multi-domain DST datasets: MultiWoz~\cite{budzianowski2018multiwoz,eric2020multiwoz} and Schema-Guided Dialogue (SGD)~\cite{rastogi2020towards}. The experimental results suggest that our proposed model, \textit{without using any DST data}, achieves a significantly higher joint goal accuracy compared to previous zero-shot DST approaches. Our contributions are summarized as the following:


\begin{itemize}
    \item We propose TransferQA, the first model that performs domain-agnostic DST \textbf{\textit{without using any DST training data}}. 
    
    \item We introduce two effective ways to construct unanswerable questions, namely, negative question sampling and context truncation, which  enable our model to handle \textit{``none''} value slots in the zero-shot DST setting;
    
    \item We demonstrate the effectiveness of our approach in two large multi-domain DST datasets. Our model achieves 1) the state-of-the-art zero-shot and few-shot results on MultiWoz and 2) competitive performance compared to a fully trained baseline on the SGD dataset.
\end{itemize}

\begin{figure}[t]
    \centering
    \includegraphics[width=\linewidth]{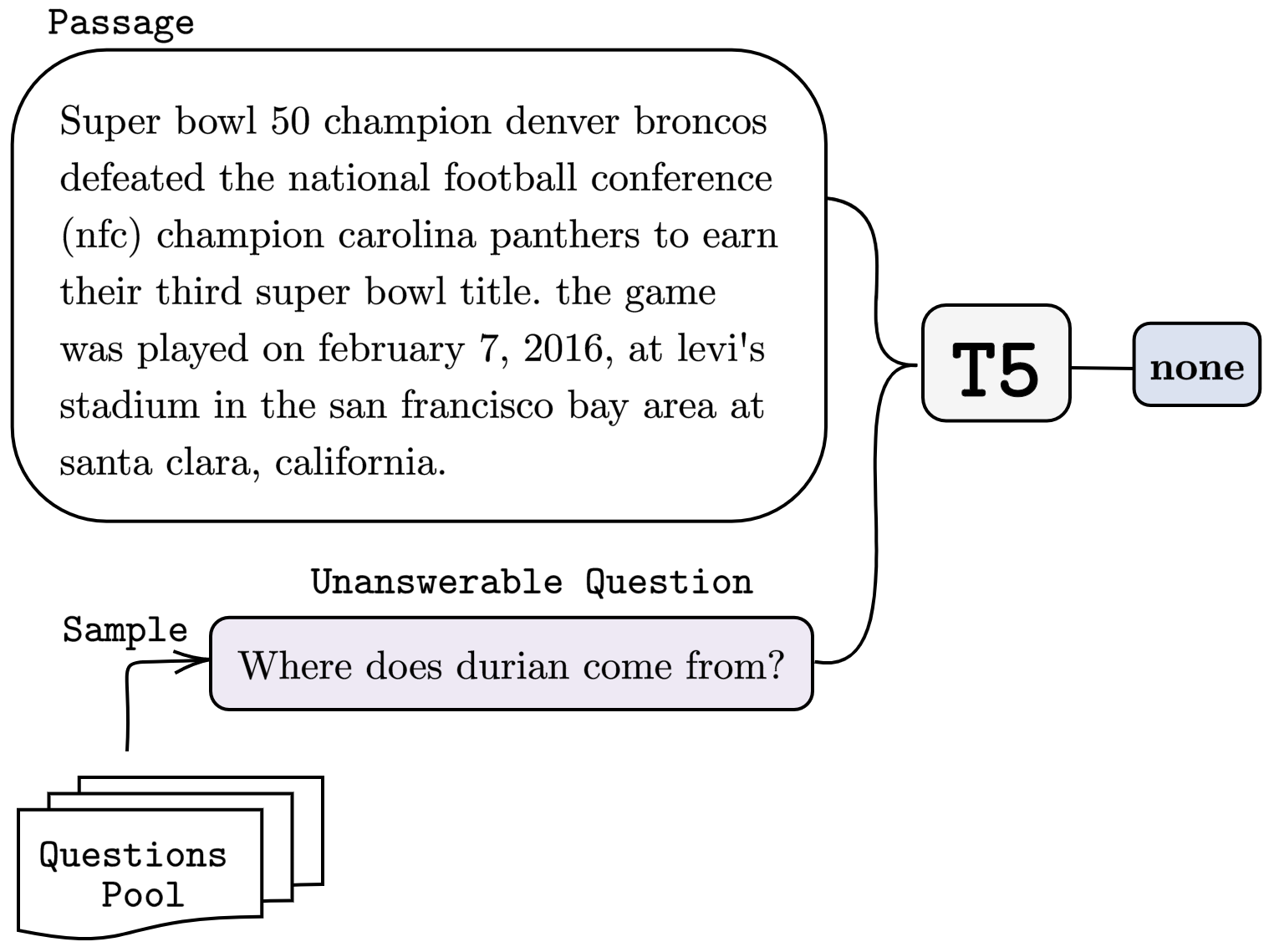}
    \caption{Negative sampling strategy for adding unanswerable questions to the training. Given a passage, we randomly sample a question from other passages and train the QA model (T5) to predict \textit{``none''}.}
    \label{fig:negative}
\end{figure}

\section{Methodology}
\subsection{Text-to-Text Transfer Learning for DST}
In multi-choice QA, each sample consists of a context passage $\mathcal{C}$, a question $q_i$, multiple answer candidates $\mathcal{A}=\{a_1,a_2, \dots,a_n\}$, and the correct answer $a_i$. In extractive QA, answer candidates are not available, and $\mathcal{A}$ become an empty set $\mathcal{A}=\emptyset$. Therefore, in QA training, the models learn to predict the answer $a_i$ to a question $q_i$ by reading the context passage $\mathcal{C}$ and answer candidates $\mathcal{A}$ (if available), while in DST inference, the models predict the value $a_i$ of a slot $q_i$ by reading the dialogue history $\mathcal{C}$ and value candidates $\mathcal{A}$ (in categorical slots).

\paragraph{QA Training.} As illustrated in Figure \ref{fig:mainfigure}, we prepend special prefixes to each input source. For instance, in multi-choice QA, \textit{``Multi-Choice Question:''} is added to the question sequence; and \textit{``Choices:''} is added to the answer candidates sequence, where each candidate is separated by a special token \textit{``[sep]''}. All the input sources are concatenated into a single sequence as input to a sequence-to-sequence (Seq2Seq) model. Then, the model generates the correct answer $a_i$ token by token. 
\begin{equation}
a_i= \mathrm{Seq2Seq}([q_i, \mathcal{A}, \mathcal{C}]).
\end{equation}
It is worth noting that some of the questions $q_i$ are unanswerable given the context. In these cases, $a_i=none$.

The training objective of our QA model is minimizing the negative log-likelihood of $a_i$ given $q_i$, $\mathcal{A}$ and $\mathcal{C}$, that is
\begin{equation}
\mathcal{L} = - \log p(a_i | q_i, \mathcal{A}, \mathcal{C}).
\end{equation}

We initialize the model parameters with T5~\cite{raffel2020exploring}, an encoder-decoder Transformer with relative position embeddings~\cite{shaw2018self}. The model is pre-trained on 750GB of clean and natural English text with a masking language modeling objective (masking out 15\% of input spans, then predicting the missing spans using the decoder).

\begin{figure}[t]
    \centering
    \includegraphics[width=\linewidth]{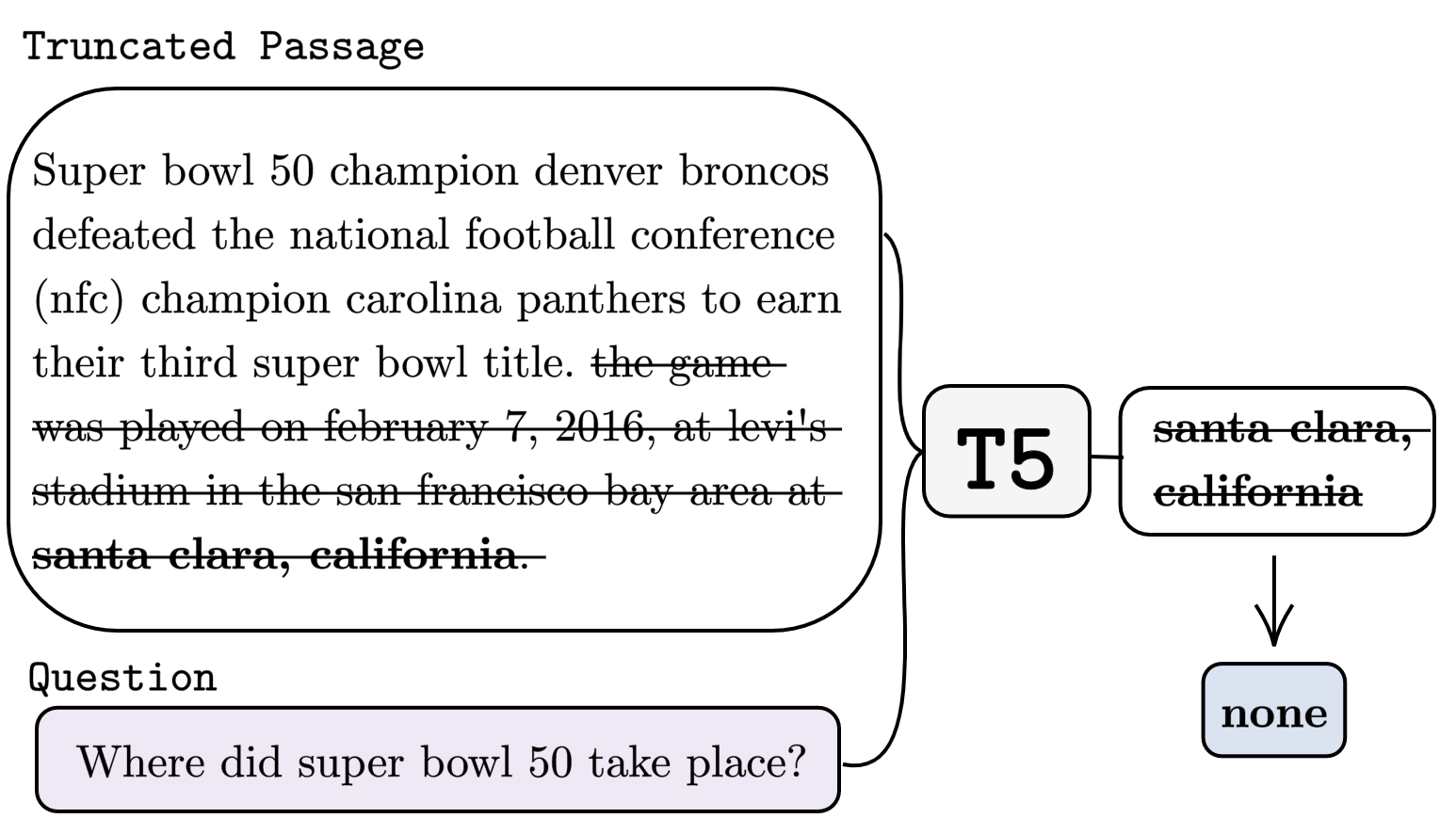}
    \caption{Context truncation strategy for generating \textit{none} values. We truncate the passage to make sure the answer span is not present in the context and thus the QA model (T5) learns to predict \textit{``none''}.}
    \label{fig:truncated}
\end{figure}

\paragraph{DST Zero-Shot.} In DST, we consider tracking a slot value as finding the answer to a slot question from a dialogue history. Therefore, we first formulate all the slots as natural language questions, with each question roughly following the format \textit{``what is the $<$slot$>$ of the $<$domain$>$ that user wants?''}. The context is a dialogue history which consists of an alternating set of utterances from two speakers, $\mathcal{C}=\{U_1,S_1, \dots, S_{t-1},U_t\}$. \textit{``user:''} and \textit{``system:''} prefixes are added to the user and system utterance, respectively. Following the QA training phase, \textit{``Multi-Choice Question:''} and \textit{``Extractive Question:''} prefixes are added to the categorical and non-categorical slot question sequence. Then, the slot question $q_i$, value candidates $\mathcal{A}$, and dialogue context $\mathcal{C}$ are concatenated into a single sequence as model input, and the model decodes the answer $a_i$ with greedy decoding.

\subsection{Unanswerable Questions} ~\label{sec:Unanswerable}
In DST, at any given turn of the conversation, the slots not mentioned by the user are marked with \textit{``none''} in the dialogue state. Especially in multi-domain dialogues, there are typically two kinds of \textit{``none''} value slots: \textit{out-of-domain} and \textit{in-domain unmentioned}. The \textit{out-of-domain} slots are the slots in other domains that are irrelevant to the current conversation. For example, when the user asks for a hotel in the center, all the slots that are not in the hotel domain (e.g., \textit{restaurant-price}) have the value \textit{``none''}. The second, \textit{in-domain unmentioned}, are those slots in the domain of interest but not yet mentioned by the user. For example, the user asks about a hotel in the center, and thus the slot \textit{hotel-star} is \textit{``none''} since the user does not specify this information. Therefore, we introduce two methods to simulate the \textit{out-of-domain} and \textit{in-domain unmentioned} slots in the QA training phase.

\paragraph{Negative Question Sampling.} The out-of-domain slots in DST are similar to out-of-context questions in QA, that is, the model must predict \textit{``none''} when the question is irrelevant to the context. To construct this kind of unanswerable question, we adapt the negative sampling strategy~\cite{mikolov2013efficient}. As illustrated in Figure~\ref{fig:negative}, during QA training, we sample these negative questions from a pool of questions collected from other passages.

\paragraph{Context Truncation.} The in-domain unmentioned slots often appear in the middle of conversations, where some of the in-domain slots have not yet mentioned by the user. We simulate such scenario by truncating the context passage from the first sentence that contains the answer span. As illustrated in Figure~\ref{fig:truncated}, given a question and a passage from a QA training set, we first truncate the passage according to the answer span annotation, then we pair the question and the truncated passage as an unanswerable sample. 

\section{Experiments}

\subsection{Datasets}

\paragraph{QA datasets.}
For the QA training, we use six extractive QA datasets such as SQuAD2.0~\cite{rajpurkar2018know}~\footnote{Note that original MRQA-2019 dataset use SQuAD~\cite{rajpurkar2016squad}, here we also add the unanswerable questions from SQuAD2.0.}, NewsQA~\cite{trischler2017newsqa}, TriviaQA~\cite{joshi2017triviaqa}, SearchQA~\cite{dunn2017searchqa}, HotpotQA~\cite{yang2018hotpotqa}, Natural Question~\cite{kwiatkowski2019natural} from MRQA-2019~\cite{fisch2019mrqa}, and two multi-choice datasets such as RACE~\cite{lai2017race} and DREAM~\cite{sun2019dream}. The main train/dev statistics are reported in Table~\ref{tab:stats}.

\paragraph{DST datasets.} The evaluation is conducted on two multi-domain task-oriented dialogue benchmark, MultiWoz~\cite{budzianowski2018multiwoz,eric2020multiwoz} and Schema-Guided-Dialogue (SGD)~\cite{rastogi2020towards}. Both datasets provide turn-level annotations of dialogue states. In MultiWoz, we follow the pre-processing and evaluation setup from \citet{wu2019transferable}, where restaurant, train, attraction, hotel, and taxi domains are used for training and testing. In SGD, the test set has 18 domains, and 5 of the domains are not presented in the training set.

\subsection{Evaluation} 
Joint Goal Accuracy (JGA) and Average Goal Accuracy (AGA) are used to evaluate our models and baselines. For JGA, the model outputs are only counted as correct when all of the predicted values exactly match the oracle values. AGA is the average accuracy of the active slots in each turn. 

In order to make consistent comparisons to the previous works on cross-domain zero-shot/few-shot DST~\cite{wu2019transferable,kumar2020ma,zhou2019multi} in MultiWoz, we compute JGA per domain as in \citet{wu2019transferable}~\footnote{\url{https://github.com/jasonwu0731/trade-dst}}. In SGD dataset, we use the official evaluation script~\footnote{\url{https://github.com/google-research/google-research/tree/master/schema_guided_dst}}.

\begin{table}[t]
\centering
\begin{tabular}{@{}rccc@{}}
\toprule
\multicolumn{1}{c}{Dataset} & Type & Train & Dev \\ \midrule
SQuAD2.0 & extractive & 130,319 & 11,873 \\
NewsQA & extractive & 74,160 & 4,212 \\
TriviaQA & extractive& 61,688 & 7,785 \\
SearchQA & extractive & 117,384 & 16,980 \\
HotpotQA & extractive & 72,928 & 5,904 \\
NaturalQA & extractive & 104,071 & 12,836 \\ 
\midrule
RACE & multiple-choice & 87,866 & 4,887 \\
DREAM & multiple-choice & 6,116 & 2,040 \\ 
\bottomrule
\end{tabular}
\caption{Datasets used in the QA pre-training. Statistics of extractive datasets (except SQuAD2.0) are taken from MRQA-2019~\cite{fisch2019mrqa}, and that of multiple-choice datasets are from RACE~\cite{lai2017race} and DREAM~\cite{sun2019dream}.}
\label{tab:stats}
\end{table}

\begin{table*}[t]
\centering
\resizebox{0.9\textwidth}{!}{
\begin{tabular}{@{}lcccccc@{}}
\toprule
\multicolumn{1}{c}{\multirow{2}{*}{Model}} & \multicolumn{6}{c}{\textbf{Joint Goal Accuracy}} \\
\multicolumn{1}{c}{} & Attraction & Hotel & Restaurant & Taxi & Train & \textbf{Average} \\ \midrule
TRADE$^\dagger$~\cite{wu2019transferable} & 20.06 & 14.20 & 12.59 & 59.21 & 22.39 & 25.69 \\
MA-DST$^\dagger$~\cite{kumar2020ma} & 22.46 & 16.28 & 13.56 & 59.27 & 22.76 & 26.87 \\
SUMBT$^{\ddagger}$~\cite{lee2019sumbt} & 22.60 & 19.80 & 16.50 & 59.50 & 22.50 & 28.18 \\
TransferQA (Ours) & \textbf{31.25} & \textbf{22.72} & \textbf{26.28} & \textbf{61.87} & \textbf{36.72} & \textbf{35.77} \\ \midrule \hspace{3mm} \textit{w/ Oracle Slot Gate} & \textit{56.81} & \textit{53.90} & \textit{56.81} & \textit{63.22} & \textit{49.57} & \textit{56.06} \\ \bottomrule
\end{tabular}
}
\caption{Zero-shot results on MultiWoz 2.1~\cite{eric2020multiwoz}. Results marked with $\dagger$ and $\ddagger$ are from \citet{kumar2020ma} and \citet{campagna2020zero}. We also report the averaged zero shot joint goal accuracy among five domains. Note that this averaged per-domain accuracy is not comparable to the JGA in full shot setting.}
\label{tab:MWOZ}
\end{table*}


\begin{table}[!t]
\centering
\resizebox{0.44\textwidth}{!}{
\begin{tabular}{@{}llcccc@{}}
\toprule
\multicolumn{2}{c}{\multirow{2}{*}{\textbf{\begin{tabular}[c]{@{}c@{}} Domain \\ \end{tabular}}}} & \multicolumn{2}{c}{SGD-baseline} & \multicolumn{2}{c}{TransferQA} \\
\multicolumn{2}{c}{} & JGA & AGA & JGA & AGA \\ \midrule
\multirow{5}{*}{\rotatebox{90}{Unseen}} & Buses* & 9.7 & 50.9 & \textbf{15.9} & \textbf{63.6} \\
 & Messaging* & 10.2 & 20.0 & \textbf{13.3} & \textbf{37.9} \\
 & Payment* & 11.5 & 34.8 & \textbf{24.7} & \textbf{60.7} \\
 & Trains* & 13.6 & 63.5 & \textbf{17.4} & \textbf{64.9} \\
 & Alarm* & 57.7 & 1.8 & \textbf{58.3} & \textbf{81.7} \\ \midrule
\multirow{13}{*}{\rotatebox{90}{Seen}} & RentalCars & 8.6 & 48.0 & \textbf{10.8} & \textbf{73.8} \\
 & Music & \textbf{15.5} & 39.9 & 8.9 & \textbf{62.4} \\
 & RideSharing & 17.0 & 50.2 & \textbf{31.2} & \textbf{61.7} \\
 & Media & 18.0 & 30.8 & \textbf{30.2} & \textbf{67.5} \\
 & Homes & 18.9 & 72.7 & \textbf{31.7} & \textbf{80.6} \\
 & Restaurants & \textbf{22.8} & 55.8 & 16.3 & \textbf{68.9} \\
 & Events & \textbf{23.5} & \textbf{57.9} & 15.6 & 56.8 \\
 & Flights & \textbf{23.9} & \textbf{65.9} & 3.59 & 42.9 \\
 & Hotels & \textbf{28.9} & 58.2 & 13.5 & \textbf{60.1} \\
 & Movies & \textbf{37.8} & \textbf{68.6} & 24.0 & 56.2 \\
 & Services & \textbf{40.9} & 72.1 & 37.2 & \textbf{75.6} \\
 & Travel & \textbf{41.5} & \textbf{57.2} & 14.0 & 24.2 \\
 & Weather & \textbf{62.0} & \textbf{76.4} & 40.3 & 59.4 \\  \midrule
\multicolumn{2}{l}{All Domain} & \textbf{25.4} & 56.0 & 20.7  & \textbf{62.2}  \\ 
\multicolumn{2}{l}{\textit{Oracle Slot Gate}} & - & - & \textit{48.0} & \textit{76.6}  \\ 
\bottomrule
\end{tabular}
}
\caption{Zero-Shot results by domain in Schema Guided Dialogue (SGD) dataset~\cite{rastogi2020towards}. The SGD-baseline is trained with the whole training set, and the results are reported by \citet{rastogi2020towards}. Domains that appear in the test set but are not present in the training set are marked with *. For TransferQA, all the domains are unseen because the model is not trained with any DST data. }
\label{tab:SGD}
\end{table}

\begin{table*}[t]
\centering
\resizebox{\textwidth}{!}{
\begin{tabular}{@{}rccccccccccccccc@{}}
\toprule
\multicolumn{1}{c}{\multirow{2}{*}{Model}} & \multicolumn{3}{c}{Hotel} & \multicolumn{3}{c}{Resturant} & \multicolumn{3}{c}{Attraction} & \multicolumn{3}{c}{Train} & \multicolumn{3}{c}{Taxi} \\
\multicolumn{1}{c}{} & 1\% & 5\% & 10\% & 1\% & 5\% & 10\% & 1\% & 5\% & 10\% & 1\% & 5\% & 10\% & 1\% & 5\% & 10\% \\ \midrule
TRADE & 19.7 & 37.4 & 41.4 & 42.4 & 55.7 & 60.9 & 35.8 & 57.5 & 63.1 & 59.8 & 69.2 & 71.1 & 63.8 & 66.5 & 70.1  \\
DSTQA & \texttt{N/A} & 50.1 & 53.6 & \texttt{N/A} & 58.9 & 64.5 & \texttt{N/A} & \textbf{70.4} & \textbf{71.6} & \texttt{N/A} & 70.3 & 74.5 & \texttt{N/A} & 70.9 & 74.1  \\
STARC & \textbf{45.9} & \textbf{52.5} & \textbf{57.3} & 51.6 & 60.4 & \textbf{64.6} & 40.3 & 65.3 & 66.2 & 65.6 & 74.1 & 75.0 & 72.5 & 75.3 & 79.6 \\
TransferQA & 43.4 & 52.1 & 55.7 & \textbf{51.7} & \textbf{60.7} & 62.9 & \textbf{52.3} & 63.5 & 68.2 & \textbf{70.1} & \textbf{75.6} & \textbf{79.0} & \textbf{75.4} & \textbf{79.2} & \textbf{80.3}  \\ \bottomrule
\end{tabular}
}
\caption{Few-shot performance on MultiWoz 2.0 in terms of Joint Goal Accuracy (JGA). \texttt{N/A} for results not presented in the original paper. All models are evaluated with 1\%, 5\%, and 10\% in-domain data. }
\label{tab:fewshot}
\end{table*}

\subsection{Implementation}
We implement TransferQA based on T5-large~\cite{raffel2020exploring}~\footnote{Source code is available in \url{https://github.com/facebookresearch/Zero-Shot-DST}}. All models are trained using the AdamW~\cite{loshchilov2018decoupled} optimizer with an initial learning rate of $0.00005$. In the QA training stage, we set the ratio of generating an unanswerable question $\alpha=0.3$, in which the ratio of negative sampled questions and truncated context is $0.95:0.05$, and we train the models with batch size 1024 for 5 epochs. 

In the DST zero-shot testing, we first treat all the slots as non-categorical and generate all the slot values. The slots that have no \textit{``none''} values are considered as active slots. Then the model generates the value of active categorical slots by using a multi-choice QA formulation. In SGD, we follow the split of non-categorical and categorical slots in the dataset, while in MultiWoz, we follow the split of MultiWoz2.2~\cite{zang2020multiwoz}, except that all the number-type slots are considered as non-categorical slots. We also apply the canonicalization technique proposed by \citet{gao2020machine} in MultiWoz.

For the few-shot experiments, the QA pre-trained models are fine-tuned with 1\%, 5\% and 10\% of the target domain data for 20 epochs. Other hyper-parameters are the same as in the QA training. We use 8 Tesla V100 GPUs for all of our experiments.

\subsection{Baselines}
\paragraph{TRADE.} Transferable dialogue state generator~\cite{wu2019transferable}, which utilizes a copy mechanism to facilitate domain knowledge transfer.

\paragraph{SUMBT.} Slot-utterance matching belief tracker~\cite{lee2019sumbt} based on the language model BERT~\cite{devlin2018bert}.

\paragraph{SGD-baseline.} A schema-guided approach~\cite{rastogi2020towards}, which uses a single BERT~\cite{devlin2018bert} model and schema descriptions to jointly predict the intent and dialogue state of unseen domains.

\paragraph{MA-DST.} A multi-attention model~\cite{kumar2020ma} which encodes the conversation history and slot semantics by using attention mechanisms at multiple granularities.

\paragraph{DSTQA.} Dialogue state tracking via question answering over the ontology graph~\cite{zhou2019multi}.

\paragraph{STARC.} Applying two machine reading comprehension models based on RoBERTa-Large~\cite{liu2019roberta} for tracking categorical and non-categorical slots~\cite{gao2020machine}.

\section{Results}
\subsection{Zero-Shot}
In Table~\ref{tab:MWOZ}, three of the baselines, TRADE, MA-DST and SUMBT, are evaluated in the cross-domain setting, where the models are trained on the four domains in MultiWoz then zero-shot on the held-out domain. Our TransferQA, \textit{without any DST training data}, achieves significantly higher JGA (7.59\% on average) compared to the previous zero-shot results. Table~\ref{tab:SGD} summarizes the results on SGD dataset, where the SGD-baseline~\cite{rastogi2020towards} is trained with the whole SGD training set. TransferQA zero-shot performance is consistently higher in terms of JGA and AGA in the unseen domains, and competitive in seen domains. The results on both datasets shows the effectiveness of cross-task zero-shot transferring. In the cross-domain transfer scenario, despite the large amount of dialogue data, only a limited number of the slots appear in the source domain. For example, MultiWoz has 8,438 dialogues with 113,556 annotated turns, but only 30 different slots in 5 domains. Thus, cross-domain transferring requires the models generalize to new slots after being trained with fewer than 30 slots. 
By contrast, in cross-task transferring, each question in QA datasets can be considered as a slot. Therefore, a model which trained with diverse questions (around 500,000) on QA datasets is more likely to achieve better generalization. 

\subsection{Few-Shot} 
Table~\ref{tab:fewshot} shows the few-shot results ond MultiWoz 2.0~\footnote{Few shot experiments are conducted on MultiWoz 2.0 for comparing with previous works.}, where TRADE~\cite{wu2019transferable} and DSTQA~\cite{zhou2019multi} are trained on four source domain on MultiWoz then finetuned with the target domain data, while STARC~\cite{gao2020machine} and our model TransferQA are first trained on the same QA datasets then finetuned with the target domain data. We experiment with 1\%, 5\% and 10\% of the target domain data. The results show that both cross-task transferring approaches (i.e., STARC and TransferQA) outperform cross-domain transferring approaches (i.e., TRADE and DSTQA) in 4 out of 5 domains. Compared to STARC, TransferQA achieves around 1\% lower JGA in the hotel domain, but consistently higher JGA on 
other domains under different data ratio settings. Especially when only 1\% of in-domain data are available, our model outperforms STARC in most domains (except hotel) by a large margin (e.g., 11.95\% in the attraction and 4.49\% in the train domain). This significant improvement can be attributed to the generated unanswerable samples, which bridge the gap between the source data distribution and the target data distribution. 


\begin{table*}[!t]
\centering
\resizebox{\textwidth}{!}{
\begin{tabular}{@{}lcccccc|cccccc@{}}
\toprule
\multicolumn{1}{c}{\multirow{2}{*}{Model}} & \multicolumn{6}{c}{\textbf{Joint Goal Accuracy}} & \multicolumn{6}{c}{\textbf{Slot Gate Accuracy}} \\
\multicolumn{1}{c}{} & Attraction & Hotel & Restaurant & Taxi & Train & \textbf{Average} & Attraction & Hotel & Restaurant & Taxi & Train & \textbf{ALL}\\ \midrule
TransferQA-base & 28.48 & \textbf{22.75} & 20.92 & \textbf{61.16} & \textbf{31.15} & \textbf{32.89} & 60.10  & 78.41 & 76.36 & \textbf{86.06} & \textbf{85.15} & \textbf{78.22}\\
\hspace{3mm} - CT & \textbf{29.51} & 21.66 & \textbf{23.37} & 58.90 & 24.13 & 31.51 & 64.81  & \textbf{78.63} & \textbf{78.83} & 83.98 & 80.11 & 78.02\\
\hspace{3mm} - CT - NQS & 23.98 & 15.54 & 18.16 & 27.03 & 12.48 & 19.44 & \textbf{65.84}  & 75.27 & 75.86 & 71.70 & 74.00 & 73.94 \\\midrule
TransferQA-large & 31.25 & \textbf{22.72} & 26.28 & 61.87 & \textbf{36.72} & \textbf{35.77} & 60.62 & 77.84 & 81.73 & 86.48 & \textbf{87.21} & 79.95\\
\hspace{3mm} - CT & \textbf{32.47} & 22.69 & \textbf{27.71} & \textbf{62.96} & 32.17 & 35.60 & 66.99 & \textbf{79.56} & \textbf{82.72} & \textbf{88.88} & 86.79 & \textbf{81.48}\\
\hspace{3mm} - CT - NQS & 24.69 & 16.22 & 23.01 & 31.54 & 23.05 & 23.70 & \textbf{69.34} & 74.82 & 80.04 & 78.87 & 83.45 & 77.95\\ \bottomrule
\end{tabular}
}
\caption{Ablation study on the effectiveness of two unanswerable question generation strategies: Context Truncation (CT) and Negative Question Sampling (NQS). The experiments are conducted on MultiWoz 2.1 with different model size. Slot Gate Accuracy measures how well the model can classify unanswerable slots. }
\label{tab:ablation}
\end{table*}

\section{Analysis}
\subsection{Impact of Unanswerable Questions} 
In Table~\ref{tab:ablation}, we study the effect of the two unanswerable question generation strategies Context Truncation (CT) and Negative Question Sampling (NQS) described in Section~\ref{sec:Unanswerable}. Applying both CT and NQS gives the best result in terms of average JGA for both TransferQA-large and TransferQA-base. While removing the CT strategy during the QA training only affects the performance in the train domain, removing both NQS and CT decreases the JGA dramatically in all the domains. This is due to the ratio of unanswerable (none) slots in MultiWoz is high (55.25\%), and removing the simulated unanswerable questions during QA training affects the Slot Gate Accuracy (SGA) in DST inference. Indeed, by adding NQS and CT, we observed large JGA improvement (around 30\%) in the taxi domain which has highest unanswerable slots ratio (71.85\%), and relatively small JGA improvement (around 10\%) in attraction and train domains where the ratios of unanswerable slots are 47.70\% and 49.58\%. Overall, these results demonstrate the importance of generating unanswerable questions.

\begin{figure}
    \centering
    \includegraphics[width=\linewidth]{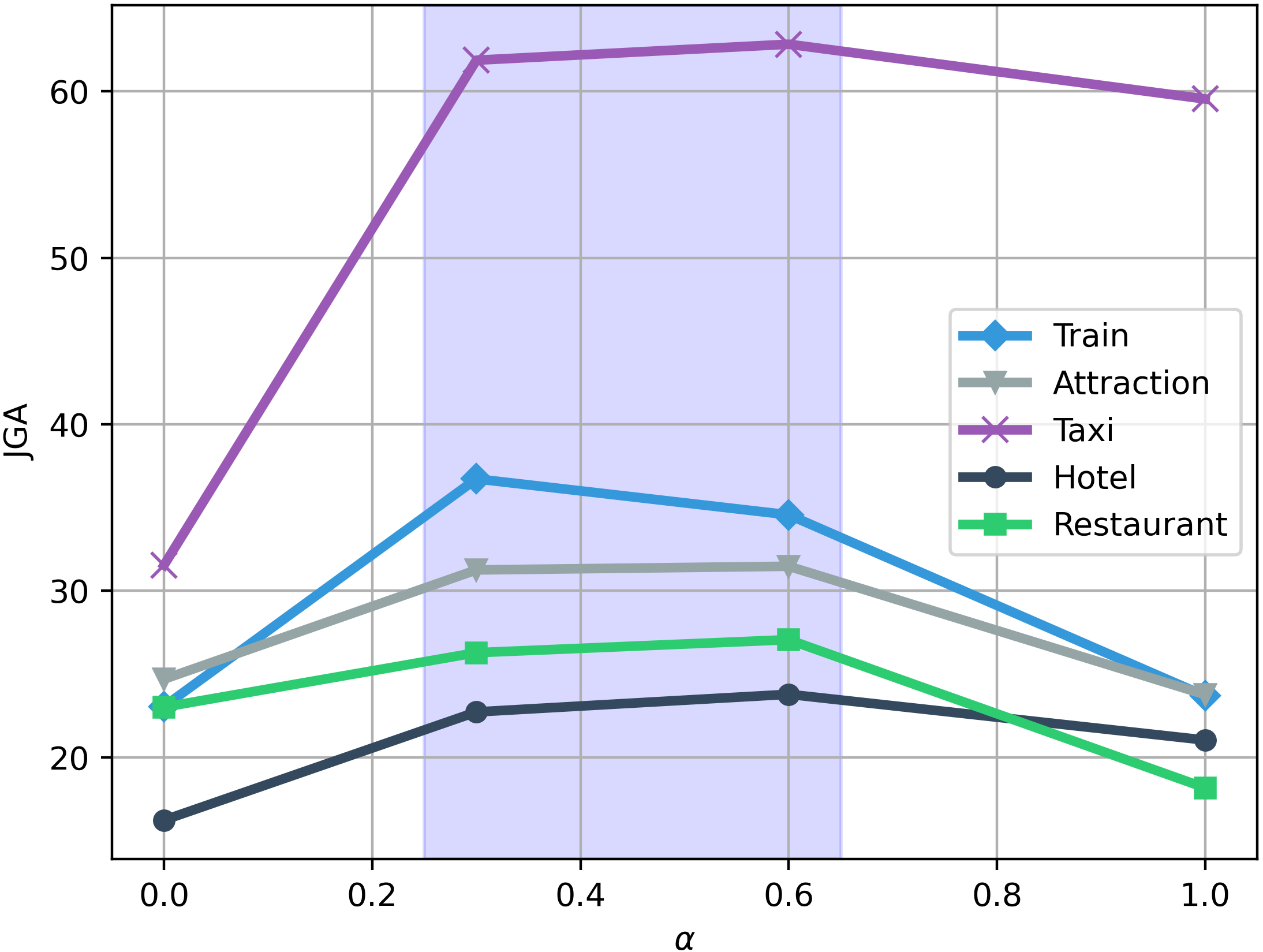}
    \caption{Joint Goal Accuracy (JGA) w.r.t. the probability of generating unanswerable questions $\alpha$. Highlighted a region of ratio where the model achieves the highest JGA.}
    \label{fig:JGAvsRatio}
\end{figure}

In Figure~\ref{fig:JGAvsRatio}, we show the effect of using different ratios $\alpha$ for generating unanswerable questions, while when it is too low, the model is not able to capture the unmentioned slots; when the ratio of unanswerable questions is too high, the model tends to over-predict \textit{``none''}. In general, we find that $\alpha=0.3$ and $\alpha=0.6$ gives the highest JGA.

\subsection{Error Analysis} 
To understand the current limitation of cross-task transfer learning, we conducted an error analysis on the results of MultiWoz 2.1 zero-shot. We found that 79.79\% of the errors come from the slot gate prediction (i.e., whether the slot is unanswerable or answerable), of which 37.54\% are false positive errors (i.e., the slot is unanswerable and the model predict answerable), 42.25\% are false negative errors (i.e., the slot is answerable and the model predicts unanswerable), and only 20.21\% of errors come from wrong value predictions of answerable slots. In Table~\ref{tab:error}, we show two typical errors that we found in the zero-shot DST setting. The first, as shown in the example of dialogue \textit{MUL2321}, is the model predicting slot values that have not been confirmed by the user yet (e.g., pricerange="expensive" etc.). The second error, as shown in dialogue \textit{PMUL0089}, is the model not capturing slot values when the user does not explicitly mention the domain (e.g., a place to stay refers to the hotel domain). These errors occurred because of question-context mismatching, and they might be addressed with well designed or leaned slot questions~\cite{li2021prefix,wallace2019universal}. We leave this exploration to the future work.

\subsection{Oracle Study} 
We further conducted an oracle study on our model by providing the gold slot gate information. The results are shown in the last row of Table~\ref{tab:SGD} and Table~\ref{tab:MWOZ}. We found that this oracle information dramatically increases the JGA (20.7\% $\rightarrow$ 48.0\% in SGD, 35.77\% $\rightarrow$ 56.06\% in MultiWoz). Therefore, by improving the accuracy of predicting \textit{``none''} value slots, we have the potential to increase the overall zero-shot DST performance by a large margin.

\begin{table*}[t]
\resizebox{\textwidth}{!}{
\begin{tabular}{@{}llll@{}}
\toprule
\multicolumn{4}{c}{\textbf{Dialogue History (MUL2321)}}    \\ \midrule
\multicolumn{4}{l}{\begin{tabular}[c]{@{}l@{}}\textbf{S:} yes I can. what restaurant are you looking for?\\ \textbf{U:} It is called \textbf{maharajah tandoori restaurant}.\\ \textbf{S:} I've located the maharajah tandoori restaurant for you. It serves \textbf{indian} food, it's in the \textbf{west} area and is in the \textbf{expensive} price range. \\ \textbf{U:} Can you book a table for \textbf{7} people at \textbf{12:30} on \textbf{tuesday}?\end{tabular}}  \\ \midrule
\textbf{Slots} & \textbf{Questions}  & \textbf{Gold Values}   & \textbf{Predicted Values} \\ \midrule
\begin{tabular}[c]{@{}l@{}}restaurant-book day\\ restaurant-book people\\ restaurant-book time\\ restaurant-name\\ restaurant-pricerange\\ restaurant-area\\ restaurant-food\end{tabular} & \begin{tabular}[c]{@{}l@{}}what is the day for the restaurant reservation?\\ how many people for the restaurant reservation?\\ what is the book time of the restaurant that the user is interested in?\\ what is the name of the restaurant that the user is interested in?\\ what is the price range of the restaurant that the user is interested in?\\ what is the area of the restaurant that the user is interested in?\\ what kind of food does user want to eat in restaurant?\end{tabular} & \begin{tabular}[c]{@{}l@{}}tuesday              \\ 7\\ 12:30\\ maharajah tandoori\\ none\\ none\\ none\end{tabular} &  \begin{tabular}[c]{@{}l@{}}tuesday              \\ 7         \\ 12:30               \\ maharajah tandoori\\ expensive          \\ west              \\ indian\end{tabular} \\ \midrule
\multicolumn{4}{c}{\textbf{Dialogue History (PMUL0089)}}   \\ \midrule
\multicolumn{4}{l}{\textbf{U:} Can you help me find a \textbf{cheap} place to stay in the \textbf{east} part of town?}  \\ \midrule
\textbf{Slots}   & \textbf{Questions}    & \textbf{Gold Values}   & \textbf{Predicted Values}  \\ \midrule
\begin{tabular}[c]{@{}l@{}}hotel-area\\ hotel-pricerange\end{tabular}  & \begin{tabular}[c]{@{}l@{}}what is the area of the hotel that the user wants?\\ what is the price range of the hotel or guesthouse that the user wants?\end{tabular}    & \begin{tabular}[c]{@{}l@{}}east\\ cheap\end{tabular}  & \begin{tabular}[c]{@{}l@{}}none\\ none\end{tabular}  \\ \bottomrule
\end{tabular}
}
\caption{Two typical errors of TransferQA zeroshot in MultiWoz 2.1. The first (top example) is predicting the values that not confirmed by the user yet, and the second (bottom example) is missing the values of implicit mentioned domain.}
\label{tab:error}
\end{table*}


\section{Related Work}
\paragraph{Machine Reading for Question Answering} (MRQA) is an important task for evaluating how well computer systems understand human language~\cite{fisch2019mrqa}. In MRQA, a model must answer a question by reading one or more context documents. There are mainly two types of MRQA tasks. The first is extractive QA~\cite{rajpurkar2016squad,rajpurkar2018know,trischler2017newsqa,joshi2017triviaqa,dunn2017searchqa,yang2018hotpotqa,kwiatkowski2019natural}, where the answer to each answerable question appears as a span of tokens in the passage. A popular approach for this task is to predict the start token and end token of the answer span~\cite{devlin2019bert}. The second is multi-choice QA~\cite{lai2017race,sun2019dream}, where the answer candidates are provided. In this task, classification-base models are usually applied to predict the correct candidate.

\paragraph{Dialogue State Tracking} is an essential yet challenging task in conversational AI research~\cite{williams2007partially,williams2014dialog}. 
Recent state-of-the-art models~\cite{lei2018sequicity,zhang2020task,wu2020tod,peng2020soloist,zhang2019find,kim2019efficient,lin2020mintl,Chen2020SchemaGuidedMD,heck2020trippy,mehri2020dialoglue,hosseini2020simple,yu2020score,li2020coco,madotto2020continual} trained with extensive annotated dialogue data have shown promising performance in complex multi-domain conversations~\cite{budzianowski2018multiwoz,eric2020multiwoz}. However, collecting large amounts of data for every dialogue domain is often costly and inefficient. To reduce the expense of data acquisition, zero-shot (few-shot) transfer learning has been proposed as an effective solution. \citet{wu2019transferable} adapt a copy mechanism for transferring prior knowledge of existing domains to new ones, while \citet{zhou2019multi} use the ontology graph to facilitate domain knowledge transfer. \citet{campagna2020zero} leverage the ontology and in-domain templates to generate a large amount of synthesized data for domain adaptation, and \citet{rastogi2020towards} apply schema descriptions for tracking out-of-domain slots. Despite the effectiveness of these approaches, a considerable amount of DST data are still required to cover a broad range of slot categories~\cite{gao2020machine}. 

On the other hand, \citet{gao2020machine} propose to utilize abundant QA data to overcome the data scarcity issue in DST tasks. The authors first train a classification model and a span-extraction model on multi-choice QA and extractive QA datasets independently. Then, they use the two QA models to track categorical and extractive slots. Compared to this approach, 
our method is fundamentally different in two aspects: 1) our model can effectively handle \textit{``none''} value slots (e.g., unmentioned and out-of-domain slots) in the zero-shot setting, which is important to DST performance as there are many \textit{``none''} slots in multi-domain dialogues;  2) our method provides a simple text-to-text input-output interface for tracking both categorical and extractive slots with a single generative model.

\section{Conclusion}
In this paper, we present TransferQA, a unified generative model that performs DST \textit{without using any DST training data}.
TransferQA uses the text-to-text transfer learning framework that seamlessly combines extractive QA and multi-choice QA for tracking both categorical slots and non-categorical slots. To enable our model to zero-shot \textit{``none''} value slots, we introduce two effective ways to construct unanswerable questions, i.e., negative question sampling and context truncation. The experimental results on the MultiWoz and SGD datasets demonstrate the effectiveness of our approach in both zero-shot and few-shot settings. We also show that improving the \textit{``none''} value slot accuracy has the potential to increase the overall zero-shot DST performance by a large margin, which can be explored in future work.

\bibliography{anthology}
\bibliographystyle{acl_natbib}




\end{document}